\documentclass[11pt]{article}
\usepackage[preprint]{acl}
\usepackage{times}
\usepackage{latexsym}

\usepackage[T1]{fontenc}
\usepackage[utf8]{inputenc}
\usepackage{microtype}
\usepackage{booktabs}
\usepackage{amsmath}
\usepackage{natbib}
\usepackage{adjustbox}
\usepackage{xcolor}
\usepackage{graphicx}
\usepackage{caption}
\usepackage{makecell}
\usepackage{booktabs}
\usepackage{multirow}
\usepackage{pifont}
\usepackage{authblk} 
\usepackage{makecell}
\usepackage{algpseudocode}  
\usepackage{algorithm}
\usepackage{algorithmicx}
\usepackage{amsmath}
\usepackage{amssymb}

\title{Memo-SQL: Structured Decomposition and Experience-Driven Self-Correction for Training-Free NL2SQL}
\author[1]{Zerui Yang}
\author[1]{Weichuan Wang}
\author[2]{Yanwei Xu\thanks{Corresponding author}}
\author[1]{Linqi Song\thanks{Corresponding author}}
\author[1]{Yudai Matsuda}
\author[2]{Wei Han}
\author[2]{Bo Bai}
\affil[1]{City University of Hong Kong, Hong Kong, PR China}
\affil[2]{Huawei Technologies Ltd.}
\begin{document}
\maketitle

\begin{abstract}
Existing NL2SQL systems face two critical limitations : (1) they rely on in-context learning with only correct examples, overlooking the rich signal in historical error–fix pairs that could guide more robust self-correction; and (2) test-time scaling (TTS) approaches often decompose questions arbitrarily, producing near-identical SQL candidates across runs and diminishing ensemble gains. Moreover, these methods suffer from a stark accuracy–efficiency trade-off: high performance demands excessive computation, while fast variants compromise quality. We present Memo-SQL, a training-free framework that addresses these issues through two simple ideas: structured decomposition and experience-aware self-correction. Instead of leaving decomposition to chance, we apply three clear strategies, entity-wise, hierarchical, and atomic sequential, to encourage diverse reasoning. For correction, we build a dynamic memory of both successful queries and historical error–fix pairs, and use retrieval-augmented prompting to bring relevant examples into context at inference time, no fine-tuning or external APIs required. On BIRD, Memo-SQL achieves 68.5\% execution accuracy, setting a new state of the art among open, zero-fine-tuning methods, while using over 10× fewer resources than prior TTS approaches.

\textbf{Contact}: zeruiyang2-c@my.cityu.edu.hk

\end{abstract}

\section{Introduction}
Recent progress in NL2SQL has been heavily reliant on either fine-tuning large language models (LLMs)~\cite{qin2024route, li2024codes} or invoking powerful but closed-source APIs~\cite{pourreza2024chase, liu2025xiyan}. While these approaches achieve strong performance on standard benchmarks, they suffer from clear drawbacks, most notably, an inability to incorporate dynamic feedback and poor generalization beyond static training setups. This has spurred growing interest in TTS~\cite{guan2025rstar, yang2025drugmcts}, a paradigm that enhances inference-time computation without modifying model parameters, thereby enabling fully open, training-free solutions based on publicly available LLMs. Despite its promise, current TTS methods~\cite{yuan2025mcts, lee2024mcs} for NL2SQL still suffer from several fundamental limitations that hinder both effectiveness and practicality.  

\begin{figure}[!tb]
    \begin{minipage}{\columnwidth}
        \centering
        \includegraphics[width=1\columnwidth]{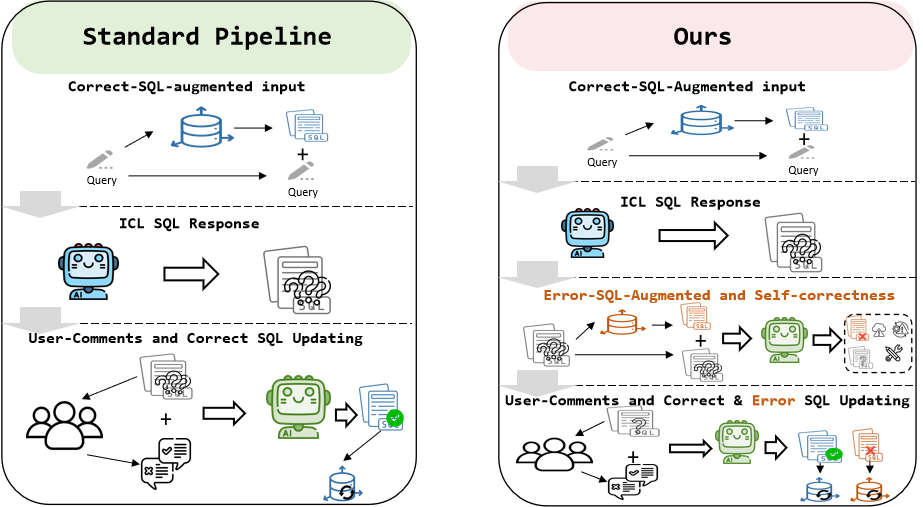}
        \caption{Conceptual comparison between a standard user-in-the-loop NL2SQL workflow and our proposed self-correction framework. In the standard approach, the model generates an initial SQL query, which is then revised through explicit user feedback, a process that typically stores only correct examples in the experience repository. In contrast, our framework introduces an automatic self-correction step after generation: it retrieves relevant error--correction pairs from a dynamic error-correction memory (including past failures) and refines the output without requiring user intervention, enabling fully training-free adaptation.}
        \label{fig:bi}
    \end{minipage}
\end{figure}

First, they often struggle to balance accuracy and efficiency, with high-performing systems incurring prohibitive computational costs~\cite{qin2024route, li2025alpha}.  

Second, many divide-and-conquer frameworks delegate the entire question decomposition process to the LLM itself~\cite{wang2023mac, pourreza2024chase}. While seemingly flexible, this approach often produces highly similar or even identical sub-question sequences across different runs~\cite{liao2025learnat}. The resulting SQL candidates are strongly correlated, which severely limits the utility of ensemble techniques like majority voting, precisely because meaningful diversity in reasoning paths is absent. Although some recent works combine divide-and-conquer with external planning modules or tool-augmented reasoning~\cite{pourreza2024chase, li2025deepeye}, our focus lies squarely on improving the core decomposition mechanism itself, without relying on auxiliary components, to unlock its full potential under a pure test-time scaling setting.

Third, existing self-correction mechanisms typically rely on static few-shot prompts that embed fixed examples of error-fix patterns~\cite{NEURIPS2023_72223cc6, xie2024decomposition}. These static demonstrations primarily serve to familiarize the model with the task format rather than genuinely enhancing its ability to diagnose and correct novel errors. Consequently, such designs cannot adapt to novel error types, evolving query formulations, or domain-specific schema quirks, often resulting in ineffective self-correction or requiring users to provide multiple rounds of clarification~\cite{huang2023large, askari2025magic, zhao2024sphinteract}. Moreover, they treat all past interactions as transient, discarding valuable feedback from user corrections, a rich source of supervision that could guide more robust and generalizable refinement.

To address these challenges, we propose Memo-SQL, a training-free TTS framework that operates exclusively with open-source LLMs and requires no external APIs or parameter updates. Memo-SQL delivers three key contributions:

\textbf{(1) A principled divide-and-conquer framework for NL2SQL.}  
Rather than leaving decomposition to the stochastic discretion of the LLM, an approach that often collapses into repetitive reasoning patterns, we formalize SQL generation as a structured divide-and-conquer process. Our framework explicitly decomposes natural language questions along multiple semantic axes (e.g., entity relations, query nesting, and operational sequencing), ensuring broad coverage of compositional structures and promoting diversity in the resulting candidate programs.

\textbf{(2) Experience-guided self-correction via error-aware in-context learning.}  
While retrieval-augmented in-context learning has been used to retrieve successful demonstrations~\cite{li2025aid, talaei2024chess}, we repurpose this mechanism for learning from failures. Specifically, we retrieve historical error-correction pairs, not just correct queries, and inject them as dynamic in-context examples during self-correction. This allows the model to explicitly reason about common failure modes and their fixes, enabling adaptive refinement that generalizes across error types and domains, without fine-tuning or handcrafted rules.

\textbf{(3) State-of-the-art performance with high efficiency.}  
Evaluated on the BIRD benchmark~\citep{li2023can}, Memo-SQL achieves 68.5\% execution accuracy on the dev-new set, setting a new state of the art among all open, zero-fine-tuning methods. Remarkably, it reduces computational overhead by over an order of magnitude compared to leading TTS approaches, demonstrating that principled test-time reasoning can simultaneously achieve high accuracy, strong adaptability, and low latency.

\begin{figure*}[!ht]
\centering
\includegraphics[width=\linewidth]{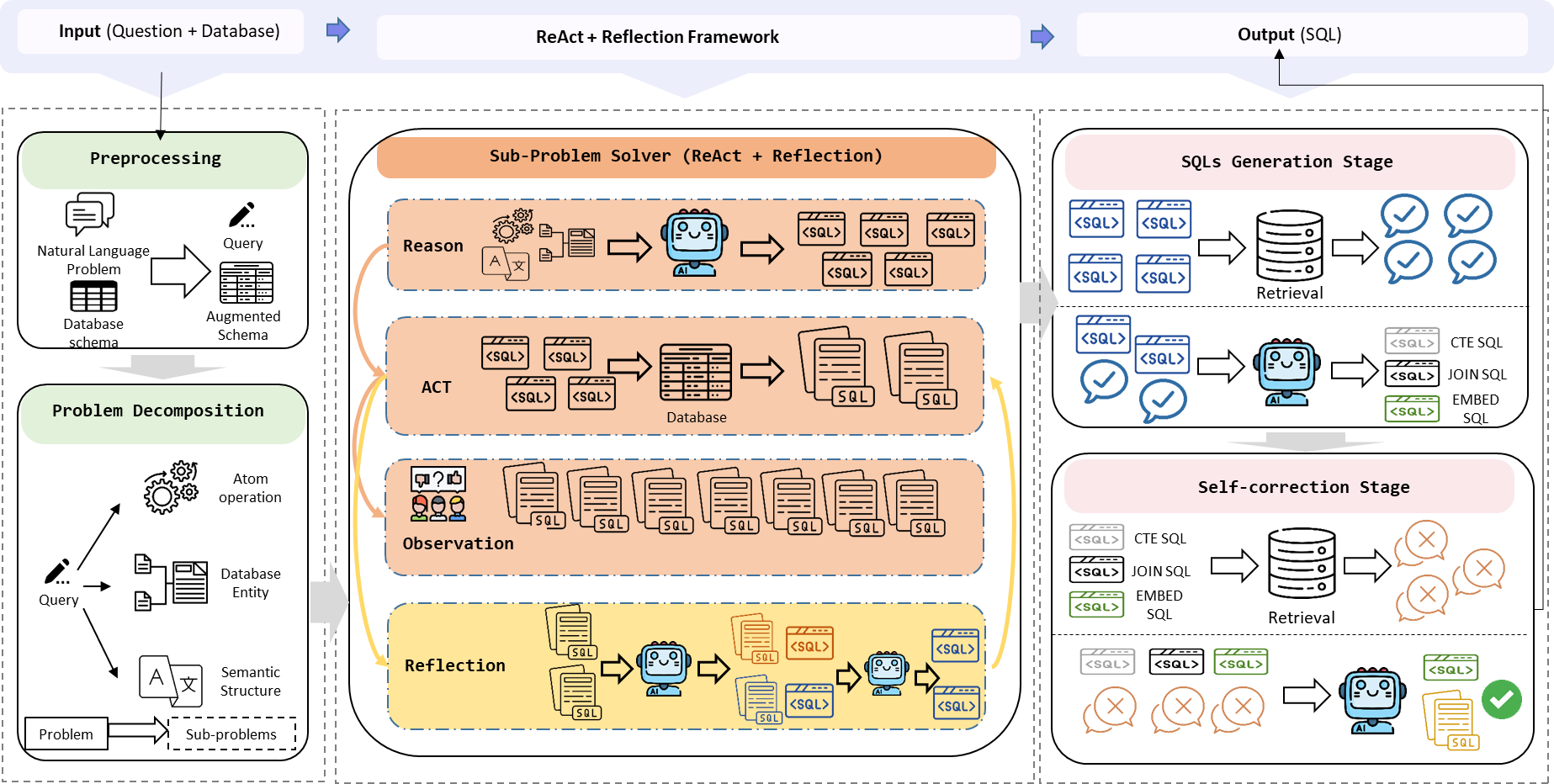}
\caption{Overview of the Memo-SQL framework, which integrates problem decomposition, ReAct+Reflection reasoning, and self-correction to generate accurate SQL queries. The pipeline begins with preprocessing and multi-strategy question decomposition. Each sub-problem is solved via an iterative ReAct+Reflection loop: (1) reasoning about semantics, (2) generating a sub-SQL query, (3) observing execution results, and (4) reflecting on potential errors to enable one-step correction. Finally, multiple SQL candidates are synthesized using few-shot in-context learning across three syntactic styles (CTE, flat JOIN, nested), followed by error-aware refinement, guided by a error-correction memory, and selection via self-consistency scoring.}
\label{fig:overview}
\end{figure*}

\section{Preliminary}

\subsection{Divide-and-Conquer for NL2SQL}
Given a natural language question $q$ and a database schema $\mathcal{S}$, the goal of NL2SQL is to generate an executable SQL query $y$. In divide-and-conquer approaches, $x$ is first decomposed into a sequence of sub-questions $\{x_1, x_2, \dots, x_k\}$, each targeting a specific aspect of $\mathcal{S}$ (e.g., entities, joins, or aggregations). The corresponding SQL fragments $\{y_1, \dots, y_k\}$ are then composed into the final query:
\begin{equation}
\begin{split}
    y &= \texttt{Compose}(y_1, y_2, \dots, y_k), \\
      &\quad \text{where } y_i = \texttt{Gen}(x_i; \theta).
\end{split}
\end{equation}
When decomposition is left entirely to the LLM (i.e., $\{x_i\}$ are sampled stochastically), the resulting $\{y_i\}$ often exhibit low structural diversity~\cite{liao2025learnat}, limiting ensemble robustness.

\subsection{In-Context Learning for Self-Correction}
Current self-correction methods use static in-context learning~\cite{NEURIPS2023_72223cc6, xie2024decomposition}. A fixed set of error–correction demonstrations $\mathcal{D}_{\text{static}} = \{(y^{\text{err}}_j, y^{\text{corr}}_j)\}_{j=1}^m$ is embedded into the prompt:
\begin{equation}
    p_{\text{prompt}} = [\mathcal{D}_{\text{static}}; \, y^{\text{err}}_{\text{new}}],
\end{equation}
and the model generates a corrected version $y^{\text{corr}}_{\text{new}} = \texttt{LLM}(p_{\text{prompt}})$. Since $\mathcal{D}_{\text{static}}$ is handcrafted and static, it cannot adapt to unseen error patterns or domain shifts.

\subsection{Experience-Aware Self-Correction via Retrieval-Augmented ICL}
As outlined in Section~\ref{sec:real_bi} and Figure~\ref{fig:bi}, real-world BI systems log both user feedback and system revisions, forming a rich repository of error–correction experiences. Inspired by this practice, we maintain a dynamic experience repository $\mathcal{M} = \{(x^{(i)}, y^{\text{err}}_{(i)}, y^{\text{corr}}_{(i)})\}_{i=1}^N$. During inference, given a newly generated (possibly incorrect) query $y^{\text{err}}$, we retrieve the top-$k$ most relevant error–correction pairs using a retriever $\mathcal{R}$:
\begin{equation}
    \mathcal{D}_{\text{retrieved}} = \mathcal{R}(y^{\text{err}}; \mathcal{M}) = \mathop{\mathrm{arg\,top\text{-}k}}_{(y^{\text{err}}_{(i)}, y^{\text{corr}}_{(i)}) \in \mathcal{M}} \text{sim}(y^{\text{err}}, y^{\text{err}}_{(i)}),
\end{equation}
where $\text{sim}(\cdot,\cdot)$ is a similarity function (e.g., embedding cosine similarity). These retrieved examples are then used as in-context demonstrations:
\begin{equation}
    y^{\text{corr}} = \texttt{LLM}\big([\mathcal{D}_{\text{retrieved}}; \, y^{\text{err}}]\big).
\end{equation}
This enables adaptive, experience-driven self-correction without fine-tuning or external APIs.

\section{Methodology}
\subsection{Overview}
The workflow of Memo-SQL comprises two phases: an offline preparation phase and an online inference phase.

In the offline phase, we construct an error-correction memory, a structured repository of quintuples $\langle q, s^{+}, s^{-}, \mathcal{E}, \delta \rangle$, each linking a natural language question $q$ to a correct SQL query $s^{+}$, an erroneous counterpart $s^{-}$, its semantic error type(s) $\mathcal{E}$, and actionable correction suggestions $\delta$.

During online inference, given a new natural language question, the system applies three complementary decomposition strategies, entity-wise, hierarchical, and atomic sequential, in parallel. For each decomposition path, the model independently executes a full ReAct+Reflect loop over its sub-questions: it (i) \textit{reasons} about the sub-task semantics, (ii) \textit{acts} by generating a sub-SQL query, (iii) \textit{observes} the execution result from the database, and (iv) \textit{reflects} on potential issues (e.g., data validity violations or semantic misalignment) based on the observed output. If reflection detects an error, the model revises the sub-query once and re-executes it to obtain a corrected intermediate result.

Because the three decomposition paths operate independently and produce complete end-to-end SQL candidates, our framework naturally implements a best-of-$N$ ($N=3$) selection scheme.

After resolving all sub-problems, the system compiles the full reasoning trace and synthesizes multiple candidate SQL queries for the original question via few-shot in-context learning. The prompt includes exemplars of $(q, s^{+})$ pairs, guiding the model to generate three syntactically diverse candidates: (1) CTE-based, (2) flat JOIN-based, and (3) nested subquery-based.

For each candidate, the system retrieves the most similar historical failure–fix instances from the error-correction memory, using a signature formed by concatenating the input question and the skeletal structure of the candidate SQL. These retrieved exemplars are injected into the prompt to enable error-aware refinement.

Finally, the system selects the best refined candidate based on a self-consistency score, which measures agreement across independently generated solutions, a proxy for robustness against decoding stochasticity.

\subsection{Offline Phase: Constructing an Error-Correction Memory}
To support error-aware refinement at inference time, we curate an error-correction memory: a structured knowledge repository that stores paired evidence of SQL failures and their fixes. An illustrative case is shown in Figure~\ref{fig:sql_error_case}.

We first define a taxonomy of nine common semantic error types (excluding pure syntax issues), covering, e.g., incorrect join constraints, missing grouping keys, and mismatched aggregation scopes (Table~\ref{tab:sql_errors}). This taxonomy serves as a controlled vocabulary to represent failure modes and enables targeted correction.

Given a training corpus (BIRD training set; \citealp{li2023can}), we elicit multiple diverse SQL candidates per question and retain instances exhibiting non-trivial semantic failures. For each such instance, we select a semantically plausible yet incorrect query as the representative erroneous query $s^-$, and pair it with the ground-truth SQL $s^+$. Conditioned on the natural language question $q$, gold SQL $s^+$, erroneous SQL $s^-$, and the database schema, a model is prompted to (i) assign one or more error types from the taxonomy and (ii) produce actionable correction suggestions. Each memory entry thus forms a structured quintuple $\langle q, s^{+}, s^{-}, \mathcal{E}, \delta \rangle$, explicitly linking an observed failure $s^{-}$ to its correction rationale $\{\mathcal{E},\delta\}$.

For efficient retrieval, each entry is encoded by concatenating the question with the skeletal structure of $s^-$ and embedding the result using GTE-base~\cite{li2023towards}. During inference, this representation enables retrieval of historically similar failure–fix patterns, which are injected as in-context evidence to guide the refinement of newly generated SQL candidates.

\begin{figure}[!tb]
    \begin{minipage}{\columnwidth}
        \centering
        \includegraphics[width=1\columnwidth]{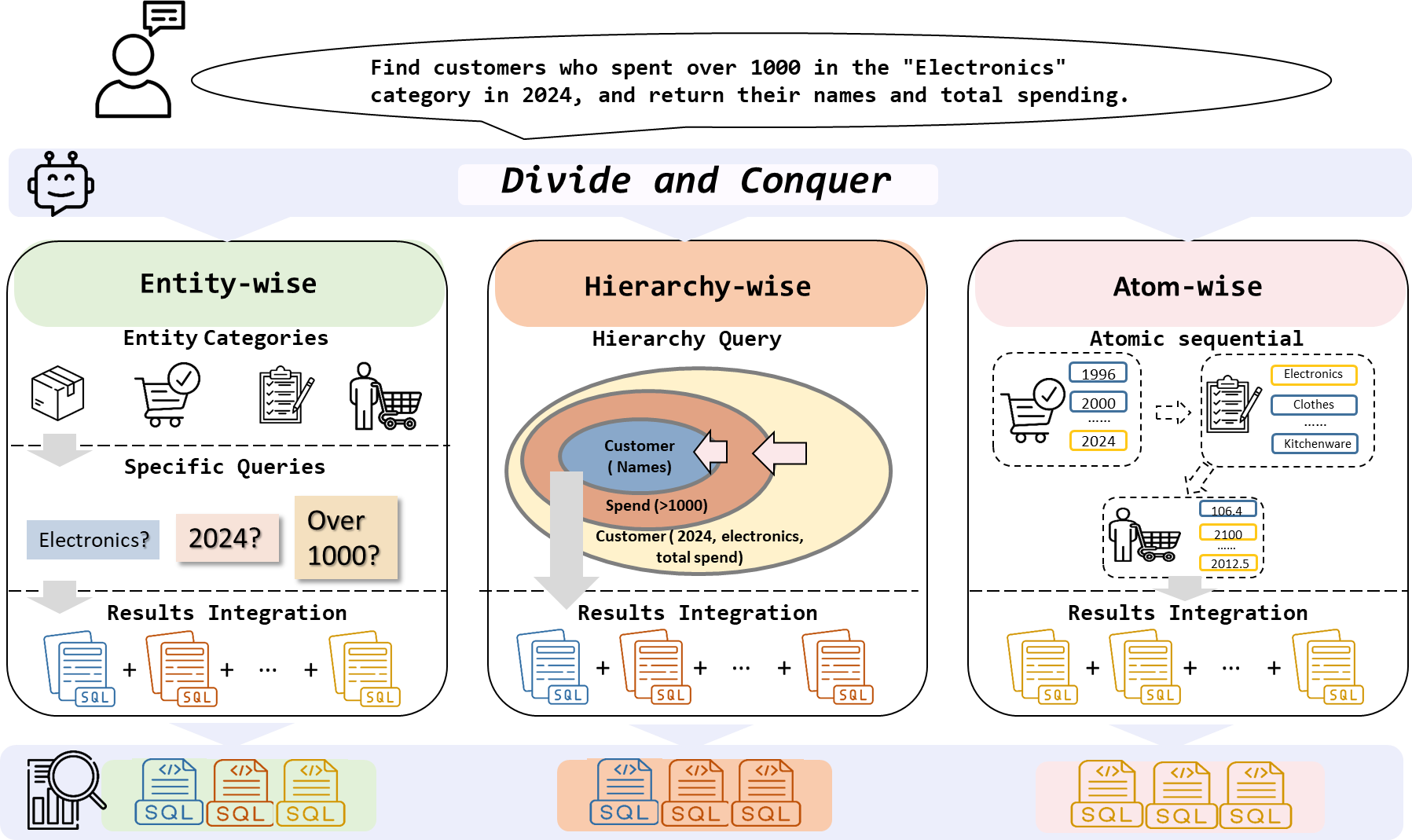}
        \caption{Illustration of the three complementary question decomposition strategies.}
        \label{fig:qd}
    \end{minipage}
\end{figure}
\subsection{Question Decomposition}
In complex natural NL2SQL tasks, accurate semantic parsing often benefits from decomposing a user question into simpler, executable sub-queries~\cite{liao2025learnat}. We identify and formalize three complementary decomposition paradigms that reflect common reasoning patterns in multi-table relational querying:

\paragraph{Table-wise Decomposition.}

Identifies all relevant tables/entities mentioned or implied in the question and generates localized sub-queries per table (with table-specific filters and projections). These are later joined via foreign keys, ideal for multi-entity conjunctive queries (e.g., “customers who bought products in New York”).

\paragraph{Hierarchical Decomposition.}

Handles nested linguistic constructs (e.g., “higher than average,” “never ordered”) by recursively decomposing the query from innermost to outermost semantics. Inner sub-queries compute aggregates or existence conditions that parameterize outer clauses, naturally aligning with SQL subquery patterns like \texttt{NOT EXISTS} or \texttt{IN}.

\paragraph{Atomic Operational Decomposition.}

Breaks the question into a sequence of atomic relational operations, selection, projection, join, grouping, and aggregation, guided by linguistic cues (e.g., “total,” “per department”). Each step maps to a valid SQL fragment, enabling incremental composition and fine-grained error localization.

Figures~\ref{fig:qd} and~\ref{fig:qd_eg} illustrate how these three decomposition strategies produce distinct reasoning paths for the same input question, highlighting their complementary roles in generating diverse and accurate SQL candidates.

\paragraph{Fallback Mechanism.}
Not all questions admit a meaningful decomposition under the above paradigms. Accordingly, we incorporate a fallback mechanism: if the model judges that a question is too simple to decompose, or that the targeted decomposition is ill-specified or would introduce unnecessary complexity, it falls back to either (i) a randomized decomposition, or (ii) no decomposition, preserving the original question for direct SQL generation.

\subsection{SQL Generation}
In this component, we adopt the ReAct+Reflect framework to generate SQL queries and proactively correct potential errors. Specifically, each sub-question is sequentially fed into the language model, which performs reasoning followed by generating a sub-SQL query (act). The generated sub-SQL is then executed to obtain intermediate results (observation). Subsequently, the model receives the current and all previously adopted strategies, and critically examines whether the current sub-SQL query contains potential issues, such as data validity violations or semantic misalignment. If errors are detected, the model revises the sub-SQL query accordingly and re-executes it. When resolving each sub-question, all prior strategies are passed to the model and incorporated into the prompt context. Finally, after all sub-problems are addressed, the model synthesizes the complete set of refined strategies, integrates them with few-shot examples, and generates the final SQL query using three distinct stylistic formulations.

\paragraph{Common Table Expressions.}

Intermediate results are materialized as named CTEs, promoting modular, stepwise reasoning and improving interpretability through explicit logical decomposition.

\paragraph{Flat JOIN-Based Formulation.}

All tables are joined in a single, flat structure using explicit \texttt{JOIN} clauses, with filters applied in \texttt{WHERE} or \texttt{ON}. This avoids nesting and aligns with classical relational algebra.

\paragraph{Nested Subqueries.}

Auxiliary logic is embedded directly within the outer query via \texttt{IN}, \texttt{EXISTS}, or scalar subqueries, enabling hierarchical representations that mirror nested natural language constructs.

\subsubsection{In-Context Learning}

In this work, ICL is employed twice to resolve ambiguities. The first instance occurs prior to final SQL generation: we concatenate the natural language question with the skeletal structure of the most recent sub-SQL query generated under the current reasoning strategy. From the database, we retrieve five semantically similar exemplars based on this combined representation and include them in the prompt as few-shot references to guide the model’s generation.

\subsection{SQL Refinement and Selection}

From the preceding stages, we obtain a total of nine candidate SQL queries, derived from three decomposition strategies combined with three distinct syntactic styles. Each candidate is subsequently subjected to rigorous semantic validation and iterative correction.

\begin{figure}[!tb]
    \begin{minipage}{\columnwidth}
        \centering
        \includegraphics[width=1\columnwidth]{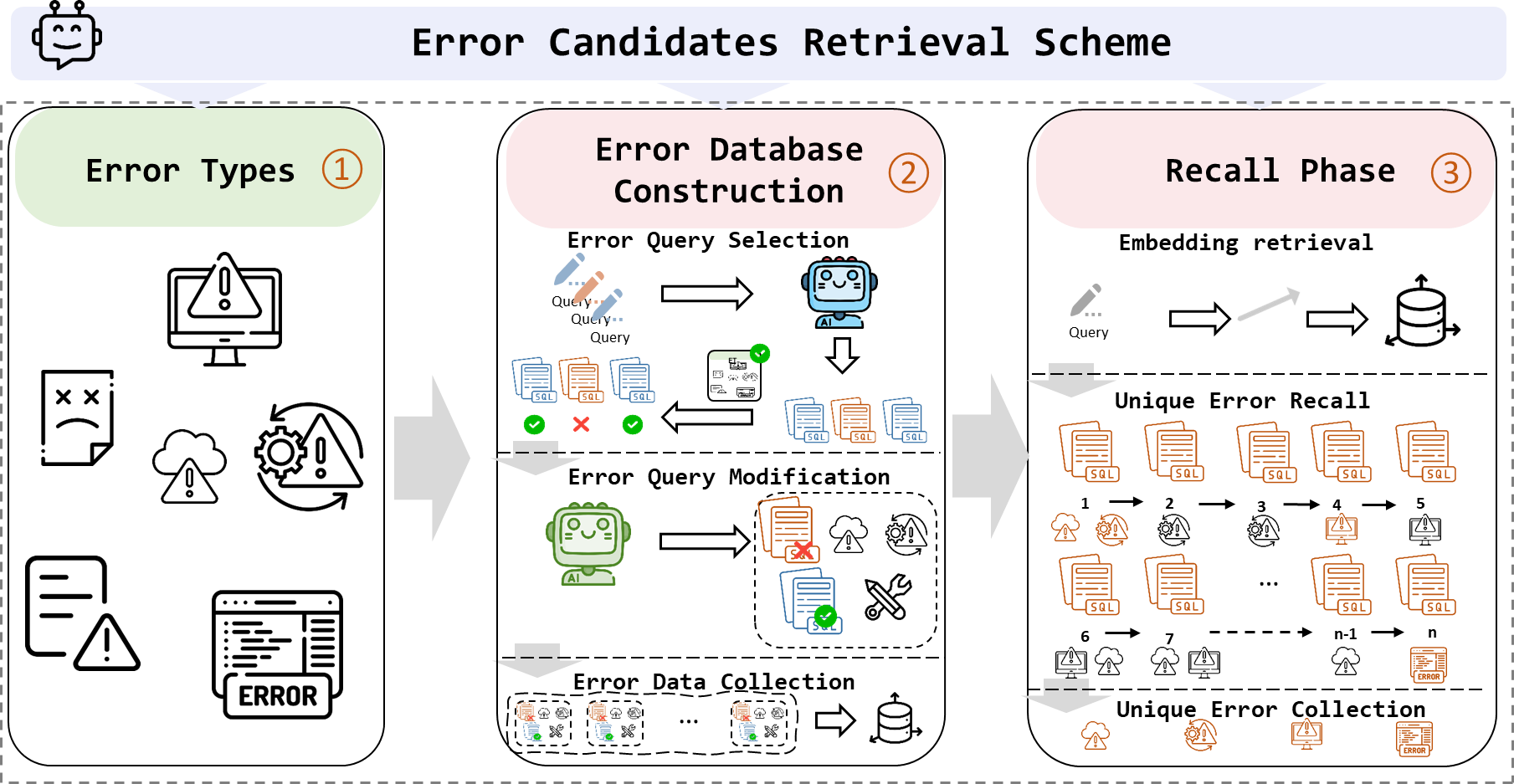}
        \caption{Construction pipeline of the error-correction memory.}
        \label{fig:bank}
    \end{minipage}
\end{figure}

\begin{table*}[h]
\centering
\small
\renewcommand{\arraystretch}{1.2}
\caption{Comparison of training-free and open-source NL2SQL methods on the BIRD dev set.}
\vspace{\baselineskip}
\begin{adjustbox}{width=\textwidth, center}
\begin{tabular}{lccccr}
\toprule
\textbf{Method} & 
\textbf{Inference Model} & 
\textbf{Selection Model} & 
\textbf{EX (\%)} \\
\midrule
% --- Group 4: No FT, open-source model (highlighted group) ---
Distillery~\citep{maamari2024death} & Llama-3.1-405B & — &  59.2 \\

ROUTE (Multi-task only)~\citep{qin2024route} & Qwen2.5-Coder-32B & Iterative Refinement &  59.8 \\

ROUTE (Multi-task only)~\citep{qin2024route} & Qwen3-Coder-30B-A3B & Iterative Refinement &  65.2 \\

Alpha-SQL$*$~\citep{li2025alpha} & Qwen2.5-Coder-32B & Majority Voting & 64.4 (64.3)$^{\dagger}$ \\

Alpha-SQL$*$~\citep{li2025alpha} & Qwen3-Coder-30B-A3B & Majority Voting & 68.2 (67.2)$^{\dagger}$\\

\textbf{Memo-SQL} & Qwen2.5-Coder-32B & Iterative Refinement & \textbf{64.9 (64.2)}$^{\dagger}$ \\

\textbf{Memo-SQL (Ours)} & Qwen3-Coder-30B-A3B & Iterative Refinement & \textbf{67.6 (68.5)}$^{\dagger}$ \\

\bottomrule
\end{tabular}
\end{adjustbox}
\footnotesize\itshape
\begin{minipage}{\textwidth}

\textbf{EX}: Execution accuracy on BIRD dev set. $*$We report results using our own evaluation script, which we argue offers a more principled and reliable assessment; see Appendix~\ref{sec:empty_results} for a detailed justification. $^{\dagger}$The results in parentheses refer to those on the BIRD dev-new set.
\end{minipage}
\label{tab:sota_comparison_bird}
\end{table*}

\subsubsection{Memory-Augmented Error Correction}

We leverage historical correction experiences as contextual guidance to steer the model toward accurate revisions. Specifically, for a given query, we first combine the original question with the skeletal form of its current SQL candidate, embed this representation, and retrieve the top-40 most similar entries from the experience bank of previously corrected examples.

A second-stage filtering is then applied: we traverse the retrieved candidates in descending order of similarity and discard any example whose set of error types is fully subsumed by those of a previously retained example. This deduplication ensures diversity in error patterns and mitigates bias toward overrepresented error categories (e.g., \texttt{SELECT} clause errors are far more common than \texttt{GROUP BY} or multi-table \texttt{JOIN} errors as shown in Figure~\ref{fig:sql_error_stat}). After this two-stage selection, each problem is typically associated with 3–5 high-quality, error-diverse reference samples.

A critic agent utilizes these samples to assess whether the current SQL contains errors. If so, it identifies the error category and provides specific revision suggestions. A refine agent then modifies the query accordingly and resubmits it to the critic for re-evaluation. This iterative refinement loop continues until either the critic confirms correctness or a maximum iteration limit is reached.

Upon completing refinement for all nine candidates, we apply majority voting, selecting the SQL query that receives the most consensus across the refined outputs as the final prediction.

\section{Experiments}

\noindent\textbf{Datasets and Evaluation Metrics.}  
We build our error-correction memory using the BIRD training set~\citep{li2023can} and evaluate our approach on four established NL2SQL benchmarks: the BIRD dev set, BIRD dev-new set~\citep{li2023can}, SPIDER development set~\citep{yu2018spider}, and CHESS-SDS dataset~\citep{talaei2024chess}. Further details are provided in Appendix~\ref{sec:data_info}.

\noindent\textbf{Implementation Details.}  
Our implementation leverages instruction-tuned models from the Qwen-Coder family, specifically Qwen2.5-Coder-32B-Instruct and the recently released Qwen3-Coder-30B-Instruct~\citep{hui2024qwen2,qwen3technicalreport}. To validate the robustness of our approach, we further conduct experiments using DeepSeek-Coder-V2-Lite (16B)~\citep{zhu2024deepseek} and Llama-3.1-8B~\citep{grattafiori2024llama3herdmodels}, both instruction-tuned variants. In the error-correction phase, we retrieve up to 40 semantically similar historical error–correction instances from our experience bank to inform refinement. The iterative correction loop is capped at three rounds to balance effectiveness against computational overhead. All results reported are averaged over three independent runs to ensure reliability.

\noindent\textbf{Baseline Methods.}  
To ensure a rigorous and reproducible comparison, we not only incorporate results reported in prior work but also re-implement two recent open-source, training-free methods based on test-time scaling, namely ROUTE~\cite{qin2024route} and Alpha-SQL~\cite{li2025alpha}, using the four aforementioned models, enabling direct comparison of both accuracy and computational cost.

Additionally, we compare our SQL correction module against three representative correction strategies employing distinct technical paradigms, which are DIN-SQL~\cite{NEURIPS2023_72223cc6}, Solid-SQL~\cite{liu-etal-2025-solid}, and SHARE~\cite{qu2025share}. More detailed descriptions of the baseline methods are provided in Appendix~\ref{sec:baseline_info}.

% \begin{table*}[t]
% \centering
% \small  % 减小字体
% \setlength{\tabcolsep}{4pt} 
% \caption{Comparison of Baseline LLMs on the CHESS-SDS Set}
% \vspace{\baselineskip}
% \label{tab:more_baseline}
% \begin{tabular}{l c c c}
% \toprule
% \textbf{Model} & \textbf{EX (\%)} & \textbf{Tokens/Query} & \textbf{Time/Query (s)} \\
% \midrule
% Deepseek-R1 & 50.3 & -- & -- \\
% GPT-4o & 53.7 & -- & -- \\
% Gemini-2.0-Flash-Thinking-Exp & 60.8 & -- & -- \\
% \midrule 
% Qwen2.5-Coder-32B     & 49.0 & 282 & 8 \\
% + ROUTE & 53.7 & 347 & 13 \\  
% + Alpha-SQL  & 58.5 & 150344 & 1839 \\
% + \textbf{Memo-SQL} & 57.6 & 7925 & 145 \\
% \midrule
% Llama-3.1-8B   & 21.1 & 332 & 4 \\
% + ROUTE & 44.2 & 414 & 9\\
% + Alpha-SQL  & 49.6 & 244775 & 1628 \\
% + \textbf{Memo-SQL} & 52.1 & 16563 & 135 \\
% \bottomrule
% \end{tabular}
% \end{table*}

\section{Results}
\textbf{BIRD dataset.} We evaluate Memo-SQL on the BIRD dev set and compare it against state-of-the-art NL2SQL methods across four categories: (1) approaches requiring fine-tuning with closed-source models, (2) training-free but closed-source methods, (3) fine-tuned open-source systems, and (4) training-free open-source frameworks.

As shown in Table~\ref{tab:sota_comparison_bird} (Full results across all method categories are reported in Table~\ref{tab:sota_comparison_bird_full} ), on the Qwen3-Coder-30B-A3B backbone, Memo-SQL achieves 67.6 on the original dev set and 68.5 on the corrected version—surpassing Alpha-SQL~\citep{li2025alpha}, which reports 68.2 (67.2) under identical evaluation conditions. While Alpha-SQL~\citep{li2025alpha} slightly outperforms our method on the original split, Memo-SQL demonstrates superior robustness on the more reliable, updated benchmark, thereby claiming the new SOTA in the open, training-free regime. Moreover, our gains are consistent across model scales: with Qwen2.5-Coder-32B, we achieve 64.9 (64.2), edging out Alpha-SQL’s~\citep{li2025alpha} 64.4 (64.3), and Distillery~\citep{maamari2024death} with Llama-3.1-405B (59.2 despite its massive scale).

Crucially, as detailed in Section~\ref{sec:efficiency_analysis}, Memo-SQL achieves this performance with significantly lower computational overhead than Alpha-SQL~\citep{li2025alpha}, striking a superior balance between effectiveness and efficiency, making it particularly suitable for real-world BI systems with latency or cost constraints.
Furthermore, Memo-SQL not only dominates the training-free open-source category but also surpasses several fine-tuned baselines, including CHESS-SQL~\citep{talaei2024chess} (61.5) and XiYan-SQL@DDL~\citep{liu2025xiyan} (62.3), despite requiring no parameter updates. This underscores the power of our experience-augmented, decomposition-driven inference framework.
The consistent improvements across diverse model architectures are further validated in Table~\ref{tab:model_algorithm_comparison}. Across all three models, ranging from 16B-scale to 32B, Memo-SQL consistently yields high execution accuracy, demonstrating strong compatibility and generalizability. 
In summary, Memo-SQL sets a new standard for open, training-free NL2SQL systems, delivering state-of-the-art accuracy on the most reliable version of BIRD, maintaining robustness across model scales, and offering favorable computational efficiency for practical deployment.

\textbf{Cross-Dataset Generalization.} To evaluate the robustness and transferability of our experience-augmented framework, we conduct experiments on the Spider dev set using an error-correction memory constructed exclusively from interactions on the BIRD training set, without any in-domain Spider data. As shown in Table~\ref{tab:sota_comparison_spider}, Memo-SQL achieves 86.5\% execution accuracy with the Qwen3-Coder-30B-A3B model, slightly below Alpha-SQL~\citep{li2025alpha} (87.8\%) but substantially outperforming ROUTE~\cite{qin2024route} (80.0\%).

When the SQL correction module is ablated (i.e., iterative refinement disabled), performance drops by 0.7\% to 85.8\%. While this degradation is less pronounced than the 2.1\% drop observed on BIRD dev set, it remains consistent with expectations: Spider queries are generally simpler, allowing strong base models to achieve high accuracy even without sophisticated correction mechanisms. Consequently, the headroom for improvement is narrower. Nevertheless, the consistent gains, despite using only out-of-domain experiences from BIRD, demonstrate that our framework effectively transfers error-correction knowledge across datasets, highlighting its strong generalization and practical applicability in real-world settings where in-domain interaction history may be unavailable.

\begin{table}[t]
\centering
\scriptsize % 更小字体（适合密集信息）
\setlength{\tabcolsep}{4pt}
\renewcommand{\arraystretch}{1.1}
\caption{Comparison of Baseline LLMs on the CHESS-SDS Set}
\vspace{\baselineskip}
\label{tab:more_baseline}
\begin{tabular}{l r r r}
\toprule
\textbf{Model} & \textbf{EX (\%)} & \textbf{Tokens/Query} & \textbf{Time/Query (s)} \\
\midrule
Deepseek-R1 & 50.3 & -- & -- \\
GPT-4o & 53.7 & -- & -- \\
\makecell[l]{Gemini-2.0-Flash-\\Thinking-Exp} & 60.8 & -- & -- \\
\midrule 
Qwen2.5-Coder-32B     & 49.0 & 282 & 8 \\
+ ROUTE & 53.7 & 347 & 13 \\  
+ Alpha-SQL  & 58.5 & 150344 & 1839 \\
+ \textbf{Memo-SQL} & 57.6 & 7925 & 145 \\
\midrule
Llama-3.1-8B   & 21.1 & 332 & 4 \\
+ ROUTE & 44.2 & 414 & 9\\
+ Alpha-SQL  & 49.6 & 244775 & 1628 \\
+ \textbf{Memo-SQL} & 52.1 & 16563 & 135 \\
\bottomrule
\end{tabular}
\footnotesize\itshape
\begin{minipage}{\columnwidth}
\end{minipage}
\end{table}

\begin{table*}[!htb]
\centering
\small
\renewcommand{\arraystretch}{1.1}
\caption{Comparison of different SQL error correction strategies on BIRD dev set. Methods are grouped into internal ablations (top) and external baselines (bottom).}
\label{tab:correction_comparison}
\begin{tabular}{@{}l l c c c c@{}}
\toprule
& \textbf{Method} & \textbf{Correction Paradigm} & \textbf{Dynamic Update} & \textbf{Training-Free} & \textbf{EX (\%)} \\
\midrule
\multicolumn{6}{c}{\textit{Internal Ablations}} \\
\addlinespace[0.5em]
& Baseline (no correction) & — & \textcolor{red}{\ding{55}} & \textcolor{green}{\ding{51}} & 66.4 \\
& + Direct Correction & Prompt Engineering & \textcolor{red}{\ding{55}} & \textcolor{green}{\ding{51}} & 66.6 \\
& + RAG (top4, unfiltered) & ICL + RAG & \textcolor{green}{\ding{51}} & \textcolor{green}{\ding{51}} & 67.5 \\
& + Random RAG (unfiltered) & ICL + RAG (random) & \textcolor{green}{\ding{51}} & \textcolor{green}{\ding{51}} & 67.6 \\
\midrule
\multicolumn{6}{c}{\textit{External SOTA Methods}} \\
\addlinespace[0.5em]
& DIN-SQL~\citep{NEURIPS2023_72223cc6} & Prompt Engineering & \textcolor{red}{\ding{55}} & \textcolor{green}{\ding{51}} & 66.5 \\
& SOLID-SQL~\citep{liu-etal-2025-solid} & ICL & \textcolor{green}{\ding{51}} & \textcolor{green}{\ding{51}} & 67.4 \\
& SHARE~\citep{qu2025share} & SFT & \textcolor{red}{\ding{55}} & \textcolor{red}{\ding{55}} & 69.2 \\
& \textbf{\underline{Memo-SQL (Ours)}} & \textbf{ICL + RAG (filtered)} & \textcolor{green}{\ding{51}} & \textcolor{green}{\ding{51}} & \textbf{68.5} \\
\bottomrule
\end{tabular}
\end{table*}

\subsection{Efficiency and Performance Trade-offs}
\label{sec:efficiency_analysis}
We evaluate Memo-SQL on the CHESS-SDS benchmark~\citep{talaei2024chess}, a challenging BIRD dev subset (Implementation details in Section~\ref{app:efficiency_setup}). As shown in Table~\ref{tab:more_baseline}, our method achieves competitive execution accuracy (EX) with dramatically lower resource use.

With Qwen2.5-Coder-32B, Memo-SQL reaches 57.6\% EX, just 0.9 points behind Alpha-SQL (58.5\%), while using 19× fewer tokens (7,925 vs. 150,344) and running 12.7× faster (145s vs. 1,839s per query). On Llama-3.1-8B, it boosts EX from 21.1\% (baseline) to 52.1\%, surpassing both ROUTE and Alpha-SQL in efficiency, with 14.8× less token usage and 12.1× lower latency than Alpha-SQL. While ROUTE~\cite{qin2024route} also uses TTS for low latency, its EX (53.7\%) lags behind ours by nearly 4 points, suggesting shallow reasoning limits its handling of complex queries.

Notably, Memo-SQL delivers strong results even on smaller models: with Llama-3.1-8B, it closes over half the gap to large proprietary systems while remaining fully open and deployable on commodity hardware. Overall, Memo-SQL offers the best balance of accuracy and efficiency, outperforming lightweight methods in correctness and heavyweight ones in speed.

\subsection{Analysis of Self-Correction Strategies}
\label{sec:correction_comparison}
As shown in Table~\ref{tab:correction_comparison}, our retrieval-augmented self-correction framework achieves 68.5\% execution accuracy, setting a new state of the art among training-free methods. In contrast, prompt engineering approaches (e.g., DIN-SQL~\citep{NEURIPS2023_72223cc6} and direct correction) yield only marginal gains over the uncorrected baseline (66.5–66.6\%), underscoring their limited capacity to handle complex semantic errors. SOLID-SQL~\citep{liu-etal-2025-solid} improves upon this by incorporating positive examples via in-context learning, reaching 67.4\%; however, it leverages only successful query demonstrations. Our method goes further by retrieving both positive and negative historical interactions, along with error-type annotations and correction rationales from the error-correction memory. This richer context enables the model to not only recognize what is correct, but also understand why a query is flawed and how to fix it, leading to more reliable and accurate revisions. Although SHARE~\cite{qu2025share} achieves slightly higher performance (69.2\%), it relies on fine-tuning three specialized models and lacks dynamic upgradability, rendering it unsuitable for interactive BI environments where systems must continuously learn from user feedback. Memo-SQL, by contrast, is fully training-free, supports real-time experience integration, and maintains strong performance without compromising deployability or privacy.

\section{Conclusion}
We presented Memo-SQL, a training-free NL2SQL framework tailored for real-world BI deployment. By combining structured divide-and-conquer decomposition with experience-driven self-correction, Memo-SQL eliminates the need for closed-source models or fine-tuning, supports dynamic updates via user feedback, and ensures data privacy. Its experience bank, populated with both correct and corrected queries, enables retrieval-augmented in-context learning that steers the model away from common semantic errors. On the BIRD benchmark, Memo-SQL achieves state-of-the-art performance among open, zero-fine-tuning methods, while reducing token usage and latency by over an order of magnitude compared to leading TTS approaches like Alpha-SQL~\citep{li2025alpha}. These advantages make Memo-SQL highly suitable for scalable, efficient, and continuously improving enterprise NL2SQL systems.

\section{Limitations}
While Memo-SQL shows strong results, it has a few practical limits. First, our experience bank is built from model-generated errors on the BIRD training set. This means the quality of self-correction depends heavily on how well those synthetic errors reflect real user mistakes, if the error types or correction patterns don't match what users actually encounter, the benefit may drop. Second, the current decomposition strategies, though structured, still rely on the LLM to correctly interpret each sub-task. On very ambiguous or poorly phrased questions, this can lead to early missteps that even later correction can't fully fix. Finally, while much faster than alternatives like Alpha-SQL, our pipeline still involves multiple LLM calls and SQL executions per query, which may be too heavy for ultra-low-latency settings (e.g., interactive dashboards requiring sub-second responses).

Nevertheless, by demonstrating that error-aware, experience-driven self-correction can work without fine-tuning or closed APIs, our approach lays a practical foundation for future open and adaptive NL2SQL systems.

\section{Ethical Statement}
This work focuses on improving the reliability and efficiency of NL2SQL systems in business intelligence (BI) settings. Our method, Memo-SQL, operates on synthetic or publicly available benchmark datasets (e.g., BIRD) and does not involve human subjects, personal data, or sensitive information. The experience repository used for self-correction is constructed from model-generated queries and simulated or anonymized user feedback; no real user interactions or private database contents are collected or stored.
While our approach enhances correctness and reduces computational cost, we acknowledge that any NL2SQL system could be misused to generate unauthorized database queries if deployed without proper access controls. Therefore, we emphasize that such systems should only be integrated into secure, permissioned environments where query execution is governed by strict authentication and authorization policies.
The authors declare no conflicts of interest.

% \clearpage 
\bibliography{ref}

\clearpage          
\appendix 
\section{Appendix}
\subsection{Real-World BI Systems}
\label{sec:real_bi}
The workflow of a real-world BI system typically involves four key components: (1) a model or a multi-agent system, (2) a generated answer, (3) user-provided feedback or revision requests, and (4) an external experience repository~\cite{negash2008business}. After receiving the generated SQL, users often inspect the execution results and propose modifications. Upon receiving such feedback, the system revises the original output. This process forms a three-party iterative loop among the model, the model output, and the user, which continues until a correct answer is produced. The final question--answer pair is then stored in a database.

When a new question is posed, the system retrieves similar historical cases from the experience repository and concatenates them into the prompt, thereby guiding the LLM to generate an answer~\cite{JoyAgent-JDGenie}.

Building on this workflow, we propose an enhanced interaction paradigm. After each round of user feedback, the system not only stores the final correct query but also captures intermediate erroneous outputs and the model’s reflections on them. Both correct and corrected–error examples are archived in the experience repository.

When a new question is posed, the system first generates an initial SQL query. A self-correction agent then retrieves historical error–correction cases whose erroneous queries are structurally or semantically similar to the current output. By analyzing these retrieved patterns, the agent assesses whether the generated SQL exhibits known failure modes. If potential issues are detected, it applies targeted edits to correct them; otherwise, the original query is retained. This transforms the conventional three-party interaction loop (model–output–user) into a four-component process: model → output → self-correction agent → user, enabling proactive error avoidance through experience-driven reasoning, without requiring retraining.

\subsection{Database Schema Preprocessing}
To facilitate accurate schema linking during SQL generation, we preprocess the structural metadata of each database, namely table names, column names, and foreign-key relationships, into a searchable semantic index.

Following established practices in NL2SQL systems~\citep{talaei2024chess, li2025alpha}, we first tokenize each schema element (e.g., ``orders.customer\_id'') and generate a compact fingerprint using MinHash~\citep{10.1145/997817.997857}. These fingerprints are precomputed for all tables and columns and stored in a local hash table.

At inference time, given a natural language question, we use a large language model to identify entity- or attribute-related keywords. We then employ locality-sensitive hashing (LSH) to efficiently retrieve candidate schema elements whose MinHash signatures are close to those of the extracted keywords. To ensure precision, retrieved candidates are further filtered using two criteria: (i) normalized edit distance below a fixed threshold, and (ii) semantic similarity above a preset cutoff, where similarity is computed as the cosine similarity between bert-large-uncased embeddings~\citep{DBLP:journals/corr/abs-1810-04805} of the schema element and the keyword.

The final set of matched schema items is incorporated into the prompt as grounded context, enabling the model to resolve lexical mismatches and correctly reference database objects during query synthesis.

\begin{table}[!ht]
\centering
\renewcommand{\arraystretch}{1.1}
\caption{Taxonomy of SQL generation errors used in our experience-augmented correction framework.}
\label{tab:sql_errors}
\begin{tabular}{l}
\toprule
\textbf{Error Type} \\
\midrule
E1: Join Logic Error \\
E2: Filter Condition Error \\
E3: Aggregation and Grouping Error \\
E4: Select Output Error \\
E5: Ordering and Limit Error \\
E6: Subquery Logical Error \\
E7: Null Handling Error \\
E8: Temporal Semantics Error \\
E9: Quantifier Intent Error \\
\bottomrule
\end{tabular}
\end{table}

\subsection{Ablation Studies}
We conduct a systematic ablation study to assess the individual contribution of each component in Memo-SQL. All experiments are evaluated on the BIRD dev-new set, and results are summarized in Table~\ref{tab:ablation}.

Starting from a strong baseline that incorporates schema preprocessing and uses Qwen3-Coder-30B as the inference engine, our full pipeline achieves an EX of 68.5\%. Removing schema linking, which grounds natural language mentions to relevant database columns and tables, leads to a modest drop of 0.8 percentage points (to 67.7\%), confirming that precise schema alignment remains beneficial even in large language model–based systems.

Replacing our structured decomposition strategy with random question decomposition reduces performance to 67.2\% (-1.3\%), highlighting the value of semantically meaningful sub-question partitioning. Similarly, ablating the ReAct+Reflect SQL generation module, a key component for enhancing query correctness through iterative reasoning and feedback loops, results in a decrease to 67.3\% (-1.2\%). This indicates that incorporating reactive adjustments and reflective refinements during SQL synthesis significantly boosts overall performance.

Disabling multi-style SQL generation (i.e., restricting output to a single syntactic form) yields 67.6\% (-0.9\%), demonstrating that generating diverse SQL formulations enhances coverage and robustness. Most notably, ablating the SQL refinement module, which performs memory-augmented error correction via iterative reflection, results in a 2.1-point drop (to 66.4\%), underscoring the critical role of post-hoc validation and correction in handling subtle semantic errors.

An unexpected finding emerges when we add ICL during the final SQL generation phase: performance decreases to 67.1\% (-1.4\%). Upon analysis, we attribute this degradation to a mismatch between the stylistic patterns in the retrieved in-context examples and our three-style generation framework. Specifically, the few-shot exemplars predominantly exhibit flat or nested SQL structures, whereas our generator deliberately produces CTE-, JOIN-, and subquery-based variants in parallel. This discrepancy appears to introduce noise or bias during decoding, leading to suboptimal outputs. We observe this degradation only with the Qwen3-Coder-30B model; for all other evaluated models, incorporating ICL during the final SQL generation phase consistently improves performance, and thus we retain this component in their respective pipelines.

Collectively, these results validate the design choices in Memo-SQL and reveal nuanced interactions between retrieval-augmented prompting and structured generation strategies.

\noindent
\begin{table}[!htb]
\centering
\small
\renewcommand{\arraystretch}{1.1}
\captionof{table}{Ablation study on BIRD-dev-new: progressive impact of each component in Memo-SQL. Reported values show absolute accuracy (\%) and changes relative to the full model (in parentheses).}
\label{tab:ablation}
\begin{tabular}{@{}l c@{}}
\toprule
\textbf{Variant} & \textbf{EX (\%)} \\
\midrule
Memo-SQL (full)                     & 68.5 \\
\quad w/o Schema Linking            & 67.7 \textcolor{gray}{(-0.8)} \\
\quad w/ Random Question Decomposition & 67.2 \textcolor{gray}{(-1.3)} \\
\quad w/o ReAct+Reflect SQL Generation & 67.3 \textcolor{gray}{(-1.2)} \\
\quad w/o Multi-Style Generation    & 67.6 \textcolor{gray}{(-0.9)} \\
\quad w/o SQL Refinement            & 66.4 \textcolor{gray}{(-2.1)} \\
\midrule
\quad + ICL in Final SQL Generation & 67.1 \textcolor{gray}{(-1.4)} \\
\bottomrule
\end{tabular}
\end{table}

\begin{table*}[!htb]
\centering
\small
\renewcommand{\arraystretch}{1.1}
\caption{Execution accuracy of different models combined with various SQL generation strategies on BIRD dev set.}
\label{tab:model_algorithm_comparison}
\begin{tabular}{@{}l c c c c@{}}
\toprule
\textbf{Model} & \textbf{Baseline} & \textbf{ROUTE} & \textbf{Alpha-SQL} & \textbf{Memo-SQL (Ours)} \\
\midrule
DeepSeek-Coder-V2-Lite & 32.5 & 51.6 & 53.4 & \textbf{53.8} \\
Qwen2.5-Coder-32B      & 52.8 & 59.8 & 64.4 & \textbf{64.9} \\
Qwen3-Coder-30B-A3B    & 62.1 & 65.2 & 68.2 & \textbf{67.6} \\
\bottomrule
\end{tabular}
\end{table*}

\subsection{On the Treatment of Empty Query Results in Evaluation}

\label{sec:empty_results}

While attempting to replicate the results of AlphaSQL, we encountered a notable inconsistency between our initial evaluation and the numbers reported in their paper. After acquiring their official evaluation script, we were able to reproduce their published scores and identified a key difference in their protocol: during majority voting, any SQL prediction that yields an empty result set is automatically discarded, and the system defaults to the next most frequent non-empty candidate.

Although this heuristic allows exact replication of their reported performance, we contend that it rests on a questionable assumption, namely, that a correct SQL query should never return an empty result. In practice, however, semantically accurate queries often produce empty outputs, especially in realistic scenarios (e.g., “Find customers who spent more than 1000 on Electronics products in 2024” in a recently launched service with limited transaction history).

To assess how common this design choice is, we inspected the public codebases and evaluation procedures of several prominent NL2SQL systems, including the official BIRD benchmark implementation~\citep{li2023can} and methods employing iterative refinement or majority voting~\cite{xie2025opensearch,NEURIPS2023_72223cc6,wang2023mac,li2024dawn}. None of these adopt explicit filtering of empty, but executable results.

In light of this, we adopt a more principled and application-aligned evaluation protocol: we only exclude queries that fail to parse or execute (i.e., those that raise runtime errors), while retaining all successfully executed queries, even if they return empty result sets. For fair comparison under consistent criteria, we re-evaluate AlphaSQL on the BIRD development set using this protocol and report the corresponding performance.

\begin{table*}[htb]
\centering
\small
\renewcommand{\arraystretch}{1.2}
\caption{Comparison of state-of-the-art NL2SQL methods on the Spider dev set.}
\begin{adjustbox}{width=\textwidth, center}
\begin{tabular}{lccccr}
\toprule
\textbf{Method} & 
\textbf{Inference Model} & 
\textbf{Selection Model} & 
\textbf{Zero-shot} & 
\textbf{Open-Source} & 
\textbf{EX (\%)} \\
\midrule
% --- Group 2: No FT, but closed-source ---
DAIL-SQL~\cite{gao2023text} & GPT-4 & Majority Voting & \textcolor{green}{Yes} & \textcolor{red}{No} & 83.6 \\
ZeroNL2SQL~\cite{fan2024combining} & GPT-4 & - & \textcolor{green}{Yes} & \textcolor{red}{No} &  84.0 \\
MAC-SQL~\cite{wang2023mac} & GPT-4 & Majority Voting & \textcolor{green}{Yes} & \textcolor{red}{No} & 86.8 \\
SuperSQL~\cite{li2024dawn} & GPT-4 & Majority Voting & \textcolor{green}{Yes} & \textcolor{red}{No} &  87.0 \\
\midrule
% --- Group 3: Need FT, but open-source ---
SFT CodeS~\cite{li2024codes} & CodeS-15B & - & \textcolor{red}{No} & \textcolor{green}{Yes} &  84.9 \\
ROUTE (Multi-task + FT)~\cite{qin2024route} & Qwen2.5-Coder-14B & Iterative Refinement & \textcolor{red}{No} & \textcolor{green}{Yes} &  87.3 \\

\midrule
% --- Group 4: No FT, open-source model (highlighted group) ---
ROUTE (Multi-task only)~\cite{qin2024route} & Qwen2.5-Coder-14B & Iterative Refinement & \textcolor{green}{Yes} & \textcolor{green}{Yes} &  80.0 \\

Alpha-SQL~\cite{li2025alpha} & Qwen3-Coder-30B-A3B-Instruct & Majority Voting & \textcolor{green}{Yes} & \textcolor{green}{Yes} & 87.8 \\

\textbf{Memo-SQL (Ours)} & Qwen3-Coder-30B-A3B-Instruct & Iterative Refinement & \textbf{\textcolor{green}{Yes}} & \textbf{\textcolor{green}{Yes}} & \textbf{86.5} \\

\quad w/o SQL Correction & Qwen3-Coder-30B-A3B-Instruct & Iterative Refinement (disabled) & \textbf{\textcolor{green}{Yes}} & \textcolor{green}{Yes} & 85.8 \\

\bottomrule
\end{tabular}
\end{adjustbox}
\vspace{6pt}
\footnotesize\itshape
\begin{minipage}{\textwidth}
\end{minipage}
\label{tab:sota_comparison_spider}
\end{table*}

\subsection{Detailed Information of Datasets}
\label{sec:data_info}
\begin{enumerate}

    \item \textbf{BIRD training set}~\citep{li2023can}: Comprising 12,751 question–SQL pairs across 95 large-scale databases, this dataset serves as the source for constructing our experience memory bank.
    
    \item \textbf{BIRD dev set}: Contains 1,534 question–SQL pairs and constitutes our primary evaluation benchmark for comparison with prior work.
    
    \item \textbf{BIRD dev-new set}: On November 6, 2025, the BIRD team released a revised version of the development set, incorporating corrections to schema annotations and SQL labels from the original release. While this updated benchmark has not yet been widely adopted—and thus lacks published baselines—we report main results on the original BIRD dev set to ensure fair comparison. Nevertheless, we provide supplementary evaluations of several representative methods on the new split to assess robustness under refined data conditions.
    
    \item \textbf{Spider dev set}~\citep{yu2018spider}: To test the generalizability of our SQL correction mechanism, we conduct cross-dataset experiments on SPIDER, which includes 1,034 question–SQL pairs drawn from heterogeneous databases spanning diverse domains.
    
    \item \textbf{CHESS-SDS dataset}~\citep{talaei2024chess}: Given that our framework operates under the TTS paradigm, where system quality depends jointly on output correctness and computational cost—we adopt a multi-dimensional evaluation protocol. In addition to execution accuracy, we measure inference efficiency via total generated tokens and end-to-end latency. These metrics are evaluated on CHESS-SDS, a high-quality, manually verified subset of the BIRD dev set that emphasizes complex queries requiring precise schema grounding and deep semantic reasoning.
\end{enumerate}

\subsection{Baseline Methods}
\label{sec:baseline_info}
\begin{itemize}
    \item \textbf{ROUTE}: A multi-agent framework that decomposes NL2SQL into four subtasks: schema linking, SQL generation, noise correction, and continuation writing. In our replication, we adopt only the agent-based orchestration without any supervised fine-tuning of the underlying LLMs.
    
    \item \textbf{Alpha-SQL}: Formulates NL2SQL as a Monte Carlo Tree Search problem, where each node corresponds to an atomic action, including question rephrasing, column-value identification, and SQL refinement, among seven predefined operations. To the best of our knowledge, Alpha-SQL represents the current state of the art among open-weight, training-free approaches.
\end{itemize}
\begin{itemize}
    \item \textbf{DIN-SQL}: Relies on prompt engineering to elicit self-verification from the model, prompting it to critique and revise its own outputs.
    
    \item \textbf{Solid-SQL}: Similar to our approach, it performs post-hoc SQL correction by retrieving database-executed examples; however, it exclusively uses correct SQL queries for retrieval and does not leverage historical error–correction pairs.
    
    \item \textbf{SHARE}: Employs SFT on three specialized modules for schema linking, logical optimization, and SQL generation. To our knowledge, SHARE currently achieves the strongest reported performance in SQL correction and serves as a strong SFT-based baseline.
\end{itemize}

Although \textbf{MAGIC}~\cite{askari2025magic} and \textbf{Sphinteract}~\cite{zhao2024sphinteract} also involve SQL correction, their primary focus lies in designing interactive prompting strategies where an external agent, assumed to have access to the ground-truth answer, guides the model toward correction. As explicitly noted in~\citet{zhao2024sphinteract}, such approaches do not address the model’s inherent inability to perform self-correction in the absence of oracle feedback. In contrast, our work aims to enable correction using only past error–fix experiences without any knowledge of the current query’s ground truth, making these methods fundamentally different in objective and thus outside the scope of our comparison.

\begin{table*}[t]
\centering
\small
\renewcommand{\arraystretch}{1.2}
\caption{Comparison of state-of-the-art NL2SQL methods on the BIRD dev set.}
\vspace{\baselineskip}
\begin{adjustbox}{width=\textwidth, center}
\begin{tabular}{lccccr}
\toprule
\textbf{Method} & 
\textbf{Inference Model} & 
\textbf{Selection Model} & 
\textbf{Training Free} & 
\textbf{Open-Source} & 
\textbf{EX (\%)} \\
\midrule
% --- Group 1: Requires fine-tuning or closed-source ---

CHESS-SQL~\citep{talaei2024chess} & Deepseek-Coder-33B & GPT-4-Turbo & \textcolor{red}{No} & \textcolor{red}{No} & 65.0 \\
Distillery~\citep{maamari2024death} & GPT-4o & — & \textcolor{red}{No} & \textcolor{red}{No} & 67.2 \\
CHASE-SQL~\citep{pourreza2024chase} & Gemini-1.5-Pro & Gemini-1.5-Flash & \textcolor{red}{No} & \textcolor{red}{No} & 73.0 \\
XiYan-SQL~\citep{liu2025xiyan} & Unreported & Unreported & \textcolor{red}{No} & \textcolor{red}{No} & 73.3 \\
\midrule
% --- Group 2: No FT, but closed-source ---
DAIL-SQL~\citep{gao2023text} & GPT-4 & Majority Voting & \textcolor{green}{Yes} & \textcolor{red}{No} & 55.9 \\
SuperSQL~\citep{li2024dawn} & GPT-4 & Majority Voting & \textcolor{green}{Yes} & \textcolor{red}{No} & 58.5 \\
MAC-SQL~\citep{wang2023mac} & GPT-4 & Iterative Refinement & \textcolor{green}{Yes} & \textcolor{red}{No} & 59.4 \\
Gen-SQL~\citep{shi2025gen} & GPT-4 & Iterative Refinement & \textcolor{green}{Yes} & \textcolor{red}{No} & 59.8 \\
RSL-SQL~\citep{cao2024rsl} & GPT-4o & Iterative Refinement & \textcolor{green}{Yes} & \textcolor{red}{No} & 67.2 \\
\midrule
% --- Group 3: Need FT, but open-source ---
DTS-SQL~\citep{pourreza2024dts} & DeepSeek-7B & — & \textcolor{red}{No} & \textcolor{green}{Yes} & 55.8 \\
SFT CodeS~\citep{li2024codes} & CodeS-15B & — & \textcolor{red}{No} & \textcolor{green}{Yes} & 58.5 \\
ROUTE (Multi-task + FT)~\citep{qin2024route} & Qwen2.5-Coder-14B & Iterative Refinement & \textcolor{red}{No} & \textcolor{green}{Yes} &  60.9 \\
CHESS-SQL~\citep{talaei2024chess} & Deepseek-Coder-33B & LLaMA3-70B & \textcolor{red}{No} & \textcolor{green}{Yes} & 61.5 \\
XiYan-SQL@DDL~\citep{liu2025xiyan} &  Qwen2.5-Coder-32B &  Qwen2.5-Coder-32B & \textcolor{red}{No} & \textcolor{green}{Yes} & 62.3 \\
XiYan-SQL@M-Schema~\citep{liu2025xiyan} &  Qwen2.5-Coder-32B &  Qwen2.5-Coder-32B & \textcolor{red}{No} & \textcolor{green}{Yes} & 67.0 \\
Reasoning-SQL~\citep{pourreza2025reasoning} &  Qwen2.5-Coder-14B & — & \textcolor{red}{No} & \textcolor{green}{Yes} & 72.3 \\
\midrule
% --- Group 4: No FT, open-source model (highlighted group) ---
Distillery~\citep{maamari2024death} & Llama-3.1-405B & — & \textcolor{green}{Yes} & \textcolor{green}{Yes} &  59.2 \\

ROUTE (Multi-task only)~\citep{qin2024route} & Qwen2.5-Coder-32B & Iterative Refinement & \textcolor{green}{Yes} & \textcolor{green}{Yes} &  59.8 \\

ROUTE (Multi-task only)~\citep{qin2024route} & Qwen3-Coder-30B-A3B & Iterative Refinement & \textcolor{green}{Yes} & \textcolor{green}{Yes} &  65.2 \\

Alpha-SQL$*$~\citep{li2025alpha} & Qwen2.5-Coder-32B & Majority Voting & \textcolor{green}{Yes} & \textcolor{green}{Yes} & 64.4 (64.3)$^{\dagger}$ \\

Alpha-SQL$*$~\citep{li2025alpha} & Qwen3-Coder-30B-A3B & Majority Voting & \textcolor{green}{Yes} & \textcolor{green}{Yes} & 68.2 (67.2)$^{\dagger}$\\

\textbf{Memo-SQL} & Qwen2.5-Coder-32B & Iterative Refinement & \textbf{\textcolor{green}{Yes}} & \textbf{\textcolor{green}{Yes}} & \textbf{64.9 (64.2)}$^{\dagger}$ \\

\textbf{Memo-SQL (Ours)} & Qwen3-Coder-30B-A3B & Iterative Refinement & \textbf{\textcolor{green}{Yes}} & \textbf{\textcolor{green}{Yes}} & \textbf{67.6 (68.5)}$^{\dagger}$ \\

\bottomrule
\end{tabular}
\end{adjustbox}
\footnotesize\itshape
\label{tab:sota_comparison_bird_full}
\end{table*}
\subsection{Experimental Setup for Efficiency Evaluation}
\label{app:efficiency_setup}

All efficiency measurements reported in Section~\ref{sec:efficiency_analysis} are obtained under identical hardware and software conditions. We deploy both Memo-SQL and baseline systems using vLLM~\citep{kwon2023efficient} on a server equipped with four GPUs, each with 24 GB of memory. Each query is processed independently (i.e., we run inference on one question at a time), and no batching is applied to ensure fair latency measurement.

For each method, we record the total wall-clock time and total number of generated tokens across all queries in the CHESS-SDS benchmark. The per-query averages are then computed by dividing these totals by the number of questions.

Importantly, we adopt different stopping criteria for Alpha-SQL and Memo-SQL to reflect their respective design philosophies:
\begin{itemize}
    \item For \textbf{Alpha-SQL}, we terminate execution immediately after all SQL candidates have been generated (as done in its original implementation), excluding the time required for majority voting or additional database executions used in final answer selection.
    \item For \textbf{Memo-SQL}, we include the full pipeline in our timing: this encompasses decomposition, multi-style candidate generation, retrieval-augmented refinement, database execution of refined candidates, and self-consistency–based majority voting. Thus, the reported latency and token counts for Memo-SQL reflect an end-to-end, deployable workflow.
\end{itemize}

\subsection{Experimental Setup for Self-Correction Ablation}
\label{app:correction_setup}

To ensure a fair comparison across correction strategies in Table~\ref{sec:correction_comparison}, we adopt a unified evaluation protocol: for every input question, we first generate nine diverse SQL candidates using the same multi-style synthesis procedure (CTE-based, flat JOIN-based, and nested subquery-based, three each). Each candidate is then independently subjected to the respective correction method, and the final prediction is selected via self-consistency majority voting over the corrected outputs.

The specific configurations for each method are as follows:

\begin{itemize}
    \item \textbf{Baseline (no correction)}: No refinement is applied; the final answer is selected by majority voting over the original nine uncorrected candidates.
    
    \item \textbf{Direct Correction}: We prompt the model to revise each candidate SQL directly, without providing any in-context examples. The prompt instructs the model to identify potential errors and output a corrected version in a single turn.
    
    \item \textbf{RAG (top4, unfiltered)}: For each candidate SQL, we retrieve the top-4 most similar entries from the error-correction memory based on the question and skeletal query structure. All retrieved entries (including erroneous queries, error types, and corrections) are concatenated into the prompt without filtering. This ablation evaluates the necessity of our two-stage retrieval-and-filtering pipeline.
    
    \item \textbf{Random RAG (unfiltered)}: Instead of similarity-based retrieval, we randomly sample 4 entries from the memory for each candidate and include them in the prompt. This tests whether performance gains stem from semantic relevance or merely from increased context length.
    
    \item \textbf{DIN-SQL}: We adapt the original prompt design of~\citet{NEURIPS2023_72223cc6}, slightly adjusting it to fit our two-agent critique–refine workflow: one agent critiques the candidate SQL, and another produces the revised version. Crucially, no historical examples are provided, only the current query and its execution feedback are used.
    
    \item \textbf{SOLID-SQL}: Following~\citet{liu-etal-2025-solid}, we retrieve the top-4 most similar correct SQL examples (question + gold query pairs) and inject them into the prompt. Like DIN-SQL, we employ a critique–refine interaction between two agents, but grounded in positive demonstrations only.
    
    \item \textbf{SHARE}: We use the official implementation and pre-trained checkpoints provided by~\citet{qu2025share} without modification.
\end{itemize}

\begin{figure*}[!ht]
\centering
\includegraphics[width=\linewidth]{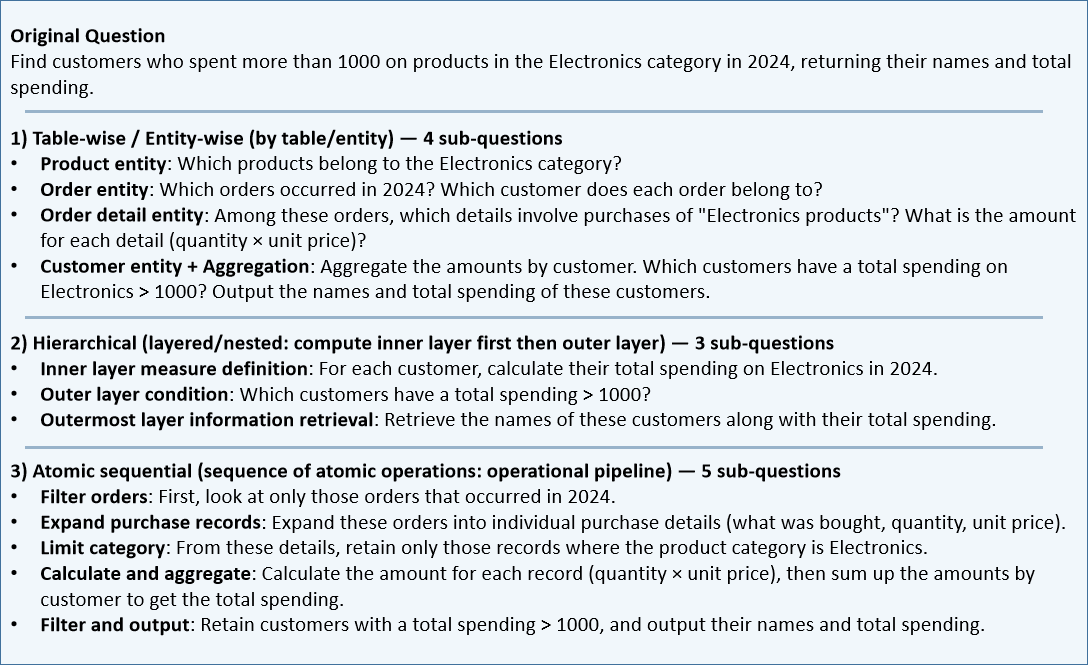}
\caption{Illustration of three complementary question decomposition strategies in Memo-SQL, demonstrated on a sample query: ``Find customers who spent more than 1000 on Electronics products in 2024.'' The methods include (1) \textit{Table-wise/Entity-wise}, decomposing the query by relevant database entities; (2) \textit{Hierarchical}, modeling nested logic through layered sub-questions; and (3) \textit{Atomic Sequential}, breaking down the query into a pipeline of fundamental relational operations. Each strategy generates a unique set of sub-questions, enabling flexible and robust reasoning for complex multi-table SQL generation.}
\label{fig:qd_eg}
\end{figure*}

\begin{figure*}[!ht]
\centering
\includegraphics[width=\linewidth]{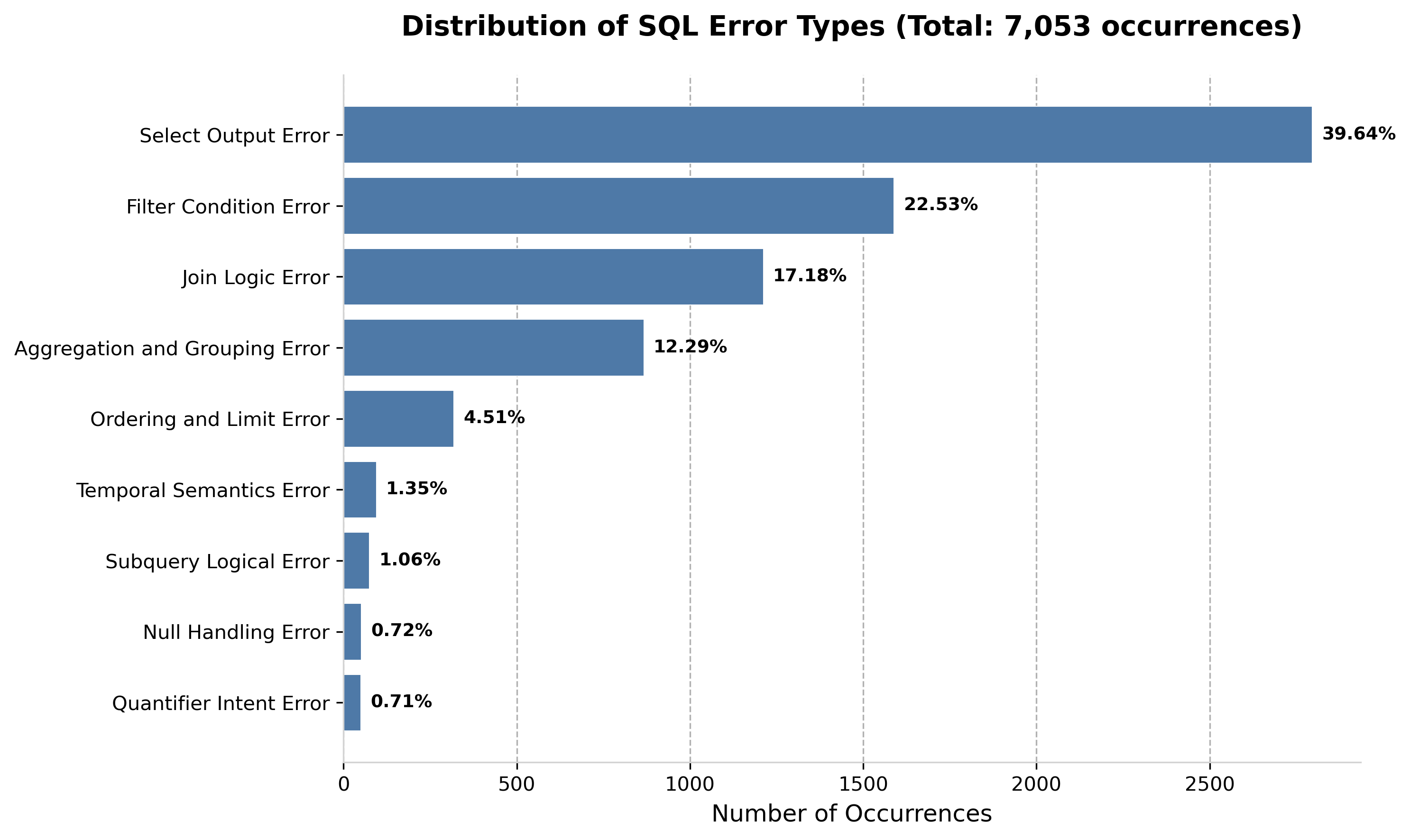}
\caption{Distribution of SQL error types across 7,053 incorrect queries generated by the Qwen3-Coder-30B-A3B model. The most frequent error category is Select Output Error (39.64\%), followed by Filter Condition Error (22.53\%) and Join Logic Error (17.18\%).}
\label{fig:sql_error_stat}
\end{figure*}

\begin{figure*}[!ht]
\centering
\includegraphics[width=\linewidth]{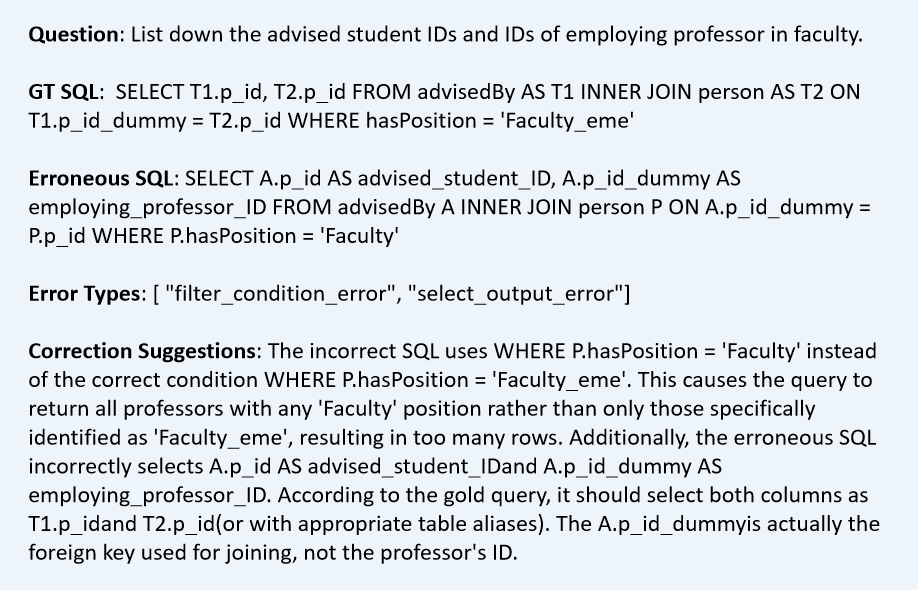}
\caption{An illustrative case of error memory in SQL generation.}
\label{fig:sql_error_case}
\end{figure*}

\subsection{Use of AI Tools}
We used AI language models solely for proofreading and polishing the language of this paper (e.g., improving grammar, clarity, and fluency). All technical ideas, methodology, experiments, analysis, and writing content were conceived and produced entirely by the authors.

\end{document}